\documentclass{article}

\PassOptionsToPackage{numbers, compress}{natbib}

\usepackage[final]{neurips_2024}
\usepackage[utf8]{inputenc} 
\usepackage[T1]{fontenc}    
\usepackage{hyperref}       
\usepackage{url}            
\usepackage{booktabs}       
\usepackage{amsfonts}       
\usepackage{nicefrac}       
\usepackage{microtype}      
\usepackage{xcolor}         
\usepackage{amsmath}
\usepackage{amssymb}
\usepackage{graphicx}
\usepackage{subfigure}
\usepackage{algorithm}
\usepackage{algorithmic}
\usepackage{multirow}
\usepackage{wrapfig}

\hypersetup{
	colorlinks=true,
	linkcolor=red,
	filecolor=blue,      
	citecolor=green,
}

\title{Classifier-guided Gradient Modulation for \\ Enhanced Multimodal Learning}

\author{Zirun Guo$^{1,2}$, Tao Jin$^{1}$\thanks{Corresponding author} , Jingyuan Chen$^{1}$, Zhou Zhao$^{1,2}$ \\ \\
$^1$ Zhejiang University, $^2$ Shanghai AI Lab\\
\texttt{zrguo.cs@gmail.com}}

\begin{document}

\maketitle

\begin{abstract}
  Multimodal learning has developed very fast in recent years. However, during the multimodal training process, the model tends to rely on only one modality based on which it could learn faster, thus leading to inadequate use of other modalities. Existing methods to balance the training process always have some limitations on the loss functions, optimizers and the number of modalities and only consider modulating the magnitude of the gradients while ignoring the directions of the gradients. To solve these problems, in this paper, we present a novel method to balance multimodal learning with \textbf{C}lassifier-\textbf{G}uided \textbf{G}radient \textbf{M}odulation (CGGM), considering both the magnitude and directions of the gradients. We conduct extensive experiments on four multimodal datasets: UPMC-Food 101, CMU-MOSI, IEMOCAP and BraTS 2021, covering classification, regression and segmentation tasks. The results show that CGGM outperforms all the baselines and other state-of-the-art methods consistently, demonstrating its effectiveness and versatility. Our code is available at \url{https://github.com/zrguo/CGGM}.
\end{abstract}

\section{Introduction}
Humans perceive the world in a multimodal way, such as sight, touch and sound. These multimodal features can provide comprehensive information to help us understand and explore the environment. Recent years have witnessed great success in multimodal learning, such as visual question answering~\citep{ben2017mutan}, multimodal sentiment analysis~\citep{tsai-etal-2019-multimodal} and multimodal retrieval~\citep{yan2024lowrank, jin2024calibrating}.

Although multimodal learning has made significant progress in recent years, inadequate use of different modality information during training remains a challenge. Theoretically, for example, \citet{wu2022characterizing} put forward the greedy learner hypothesis which states that a multimodal model learns to rely on one of the input modalities, based on which it could learn faster, and does not continue to learn to use other modalities. \citet{huang2022modality} find that during joint training, multiple modalities will compete with each other and some modalities will fail in the competition. Experimentally, on some multimodal datasets, there is little improvement in accuracy between training with only one modality and training with all modalities~\citep{tsai-etal-2019-multimodal, vielzeuf2018centralnet}. These theoretical analyses and experimental results demonstrate the inefficiency of multimodal learning to fully utilize and integrate information from different modalities.

To deal with this problem, recent studies~\citep{wu2022characterizing, peng2022balanced, li2023boosting, fan2023pmr, zhang2023provable, fu2023multimodal} investigate the training process of multimodal learning and propose gradient modulation strategies to better integrate the information of different modalities and balance the training process in some situations. However, all of these methods can not be applied easily for some limitations. For example, \citet{wu2022characterizing}, \citet{peng2022balanced}, \citet{li2023boosting} and \citet{hua2024reconboost} propose balancing methods based on cross-entropy loss for classification tasks. For regression tasks or other tasks, we can not use these strategies. Besides, most of these methods can just deal with situations where there are only two modalities. For example, \citet{wu2022characterizing} propose the conditional learning speed which is difficult to calculate and employ if there are more than two modalities. For situations where there are more modalities, these methods can not be applied. Furthermore, most of these methods only consider modulating the magnitude of the gradients while ignoring the directions of the gradients.

Based on the above observations, in this paper, we propose a novel method to balance multimodal learning with \textbf{C}lassifier-\textbf{G}uided \textbf{G}radient \textbf{M}odulation (CGGM). In CGGM, we consider a more general situation with no limitations on the type of tasks, optimizers, the number of modalities, etc. Additionally, we consider both the magnitude and directions of the gradients to fully boost the training process of multimodal learning. Specifically, we add classifiers to evaluate the utilization rate of each modality and obtain the unimodal gradients. Then, we leverage the utilization rate to adaptively modulate the magnitude of the gradients of encoders and use the unimodal gradients to instruct the model to optimize towards a better direction.

We conduct extensive experiments on four multimodal datasets: UPMC-Food 101~\citep{7169757}, CMU-MOSI~\citep{mosi}, IEMOCAP~\citep{Busso2008IEMOCAPIE}, and BraTS 2021~\citep{baid2021rsna}. UPMC-Food 101 and IEMOCAP are classification tasks, CMU-MOSI is a regression task, and BraTS 2021 is a segmentation task. CGGM outperforms all the baselines and other state-of-the-art methods, demonstrating its effectiveness and universality. In summary, our contributions are as follows:
\begin{itemize}
   \item We propose CGGM to balance multimodal learning by both considering the magnitude and direction of the gradients.
   \item CGGM can be easily applied to many multimodal tasks and networks with no limitations on the type of tasks, optimizers, the number of modalities, etc. which indicates its versatility.
   \item Our proposed CGGM brings consistent improvements to various tasks, including classification, regression and segmentation tasks. Extensive experiments show that CGGM outperforms other state-of-the-art methods, demonstrating its effectiveness.
\end{itemize}

\section{Related Work}

\noindent\textbf{Multimodal Learning.} One of the main challenges of multimodal learning is how to effectively utilize and integrate the information from different modalities to complement each other. According to the fusion strategies, there are three main multimodal fusion strategies: early fusion, intermediate fusion and late fusion. In early fusion methods~\citep{liang-etal-2018-multimodal, wang2018words}, raw data from different modalities is combined via concatenation or other methods at the input level before being fed into a model. Intermediate fusion~\citep{joze2020mmtm} methods combine data from different modalities at various intermediate processing stages within a model architecture. Late fusion~\citep{ben2017mutan, tsai-etal-2019-multimodal} methods process data from each modality independently through separate models and combine them at a later stage. In general, late fusion is the predominant method used in multimodal learning. The main reason~\citep{joze2020mmtm} is that the architecture of each unimodal stream has been carefully designed over the years to achieve state-of-the-art performance for each modality. Therefore, we can leverage these pre-trained models~\citep{devlin2018bert, dosovitskiy2020image} to achieve better results. Therefore, in this paper, our method is based on late fusion.

These fusion strategies are able to integrate information from different modalities effectively, but they have limited improvements to utilize information from different modalities to complement each other. In other words, they are not able to deal with the modality competition~\citep{huang2022modality} or imbalanced multimodal learning. When the dominant modality is missing~\citep{guo2024multimodal} or corrupted, the performance would degrade significantly. Different from these fusion strategies, our method aims to make relatively full use of the information of each modality and address the imbalanced multimodal learning.

\noindent\textbf{Balanced Multimodal Learning.} The inefficiency in fully utilizing and integrating information from multiple modalities poses a great challenge to the multimodal learning field. Some studies~\citep{tsai-etal-2019-multimodal, vielzeuf2018centralnet} present that there is little improvement in accuracy between training with only one modality and training with all modalities. \citet{wang2020makes} show that multimodal models using multiple modalities can be even inferior to those using only one modality. To balance the multimodal learning process and fully utilize different modalities, a series of balanced multimodal learning methods~\citep{wu2022characterizing, peng2022balanced, li2023boosting, fan2023pmr, zhang2023provable, fu2023multimodal, du2023uni, hua2024reconboost} are proposed. \citet{wu2022characterizing} propose the conditional learning speed to capture the relative learning speed between modalities and balance the learning process. \citet{peng2022balanced} propose a gradient modulation strategy that adaptively controls the optimization of each modality via monitoring the discrepancy of their contribution towards the learning objective. More recently, \citet{fan2023pmr} propose the prototypical modal rebalance strategy to introduce different learning strategies for different modalities. \citet{li2023boosting} propose an adaptive gradient modulation method that can boost the performance of multimodal models with various fusion strategies. \citet{hua2024reconboost} dynamically adjust the learning objective with a reconcilement regularization against competition with the historical models.

\begin{figure*}
  \begin{center}
  \centerline{\includegraphics[width=\columnwidth]{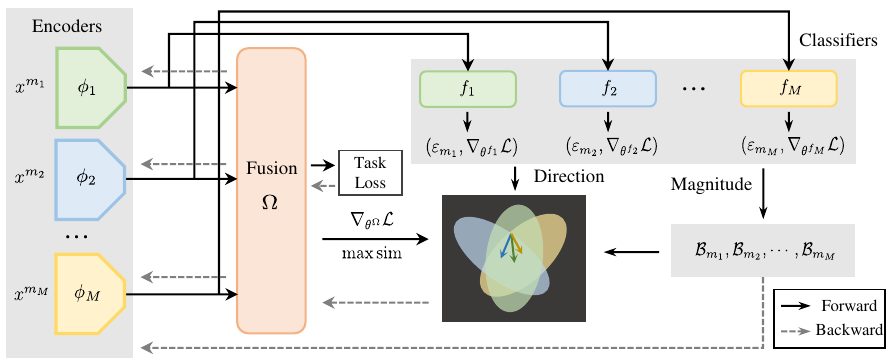}}
  \caption{The overall architecture of CGGM. During the training stage, classifiers are introduced to calculate the directions of unimodal gradients and evaluation metrics. During the inference stage, the classifiers are discarded.}
  \label{fover}
  \end{center}
  \vskip -0.2in
\end{figure*}

However, all of these previous works have certain limitations and can only be used in some specific situations. For example, \citet{wu2022characterizing} propose conditional learning speed based on intermediate fusion strategy which makes it hard to apply to situations where there are more than two modalities or where the network is not based on intermediate fusion. \citet{peng2022balanced}, \citet{fan2023pmr}, \citet{fu2023multimodal}, \citet{li2023boosting} and \citet{hua2024reconboost} propose the balancing strategies with the assumption of the cross-entropy loss function mainly for classification. Particularly, \citet{peng2022balanced} employ the SGD optimizer. Different from these methods, we consider a more general situation with no limitations on the number of modalities, the optimizer, the loss function and so on. Additionally, most of existing methods only consider the magnitude of the gradients and ignore the directions of the gradients. In contrast, we consider both of them.

\section{Proposed Method}

\subsection{Problem Settings}\label{s31}
Suppose there are $M$ modalities, referred to as $m_1,$ $m_2,$ $\cdots, m_M$. We denote the multimodal dataset as $\mathcal D=$ $\{(\boldsymbol{x_i}, y_i)\}_{i=1}^N$, where $N$ is the number of data in the dataset and $\boldsymbol{x_i}=(x_i^{m_1}, x_i^{m_2}, \cdots, x_i^{m_M})$.

We consider the most common structure (Figure~\ref{fover}) in multimodal models, where the inputs of different modalities are first fed into modality-specific encoders and then the representations of all modalities are inputted into a fusion module. We denote the encoder of modality $m_i$ as $\phi_i$ where $i=1,2,\cdots, M$ and the fusion module as $\Omega$.

\begin{figure}
  \vskip 0.2in
  \begin{center}
  \subfigure[]{\includegraphics[width=.32\columnwidth]{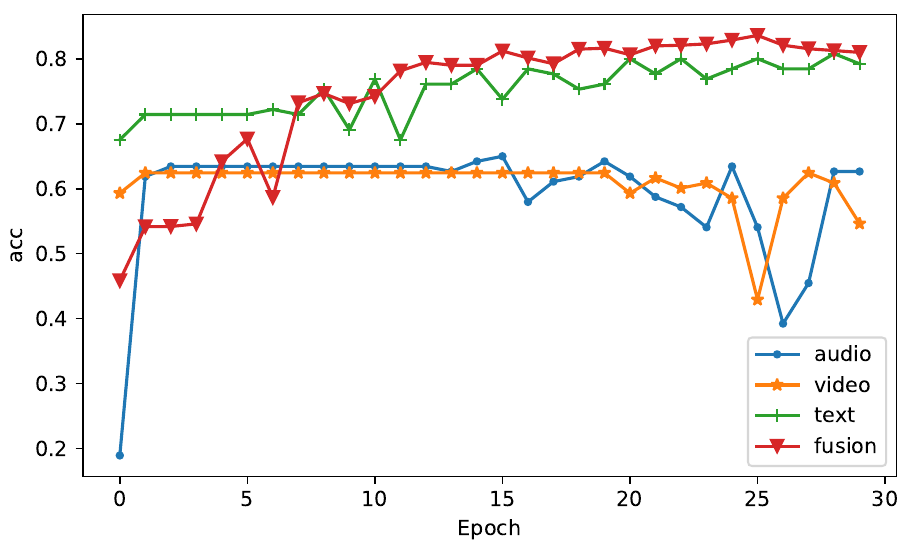}}\hspace{5pt}
 \subfigure[]{\includegraphics[width=.32\columnwidth]{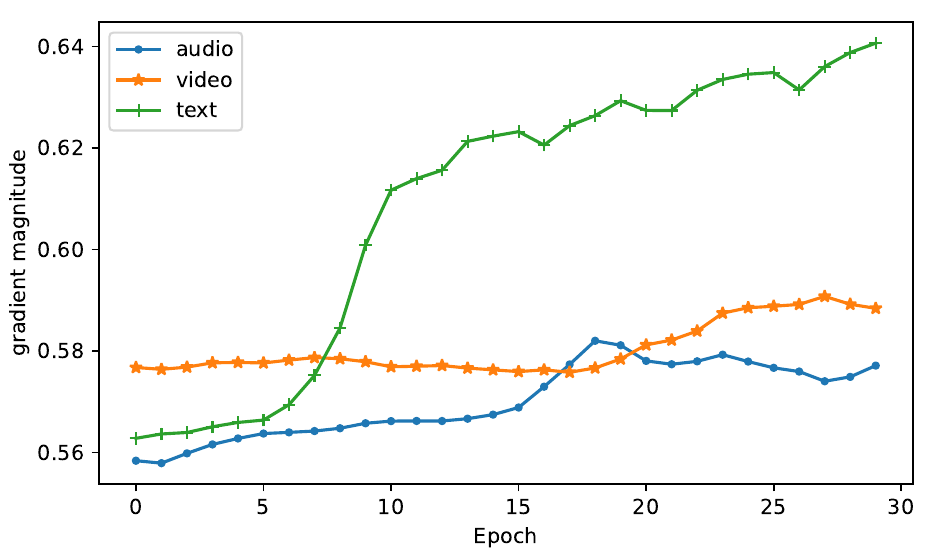}}\hspace{5pt}
 \subfigure[]{\includegraphics[width=.32\columnwidth]{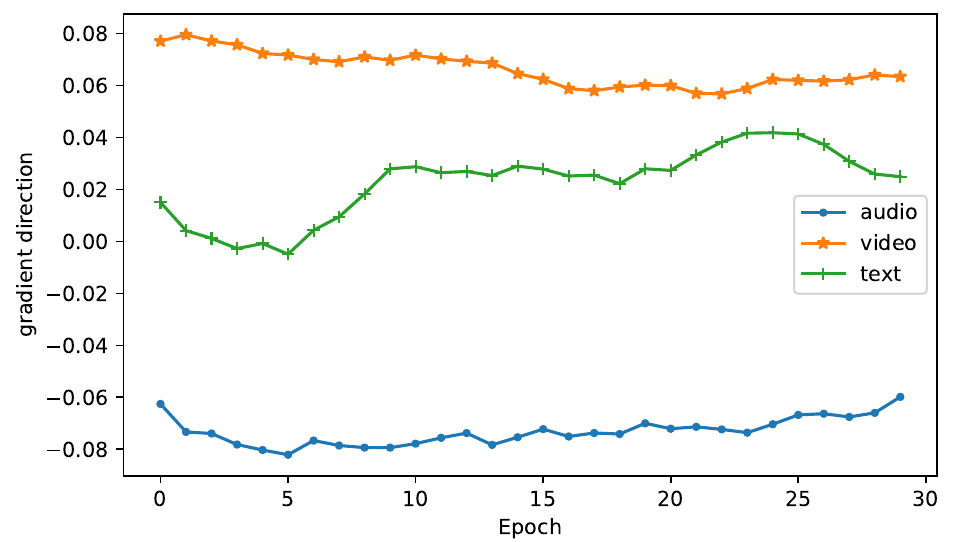}}\hspace{5pt}
\\
  \caption{(a) Accuracy of each modality and the fusion. (b) Gradient magnitude of each modality. We use the Euclidean norm of the gradient vector to represent the gradient magnitude. (c) Gradient direction between each modality and their fusion. We use cosine similarity to represent the direction between two gradient vectors. We get all the results on the CMU-MOSI dataset.}
  \label{inifig}
  \end{center}
  \vskip -0.2in
\end{figure}

For the forward propagation, the features are first inputted into the encoder:
\begin{equation}\label{e1}
  h_i=\phi_i(x^{m_i})
\end{equation}
where $h_i$ is the representation of modality $m_i$. After obtaining the representations of all modalities, the fusion module is applied:
\begin{equation}\label{e2}
  \hat y=\mathcal F(\Omega([h_1,h_2,\cdots,h_M]))
\end{equation}
where $\hat y$ is the prediction, $[\cdots]$ is the concatenation operation, and $\mathcal F(\cdot)$ is the prediction head to predict the answer. $\Omega(\cdot)$ fuses the multimodal representations and outputs the fused feature as the prediction token.

\subsection{Gradient Analysis}\label{s32}
To introduce CGGM, we first analyze the gradient updating process. We denote the loss function as $\mathcal L(\theta)=\frac{1}{N}\sum_{i=1}^N\ell (\hat y_\theta^i, y^i)$ where $\theta$ represents the parameters of the network, $\hat y_\theta^i$ is the prediction and $y^i$ is the ground truth. For simplicity, we use $\hat y^i$ to represent the predictions in the following context. Different from previous methods which only consider cross-entropy loss~\citep{peng2022balanced, fan2023pmr,hua2024reconboost}, our $\mathcal L$ can be cross-entropy loss, L1 loss or any other loss functions. With the \textit{Gradient Descent} (GD) optimization method, the parameters of the fusion module $\Omega$ and encoders $\phi_i$ can be updated as:
\begin{equation}\label{e3}
  \begin{aligned}
  \theta_{t+1}^{\Omega}&=\theta_{t}^{\Omega}-\alpha\nabla_{\theta^{\Omega}}\mathcal L(\theta_{t}^{\Omega}) =\theta_{t}^{\Omega}-\alpha\frac{1}{N}\sum_{n=1}^N\left (\frac{\partial\mathcal F}{\partial\Omega}\right )^\top\frac{\partial\ell\left (\hat y^n, y^n\right )}{\partial\mathcal F}
  \end{aligned}
\end{equation}
\begin{equation}\label{e4}
  \begin{aligned}
  \theta_{t+1}^{\phi_i}&=\theta_{t}^{\phi_i}-\alpha\nabla_{\theta^{\phi_i}}\mathcal L(\theta_{t}^{\phi_i}) =\theta_{t}^{\phi_i}-\alpha\frac{1}{N}\sum_{n=1}^N\left (\frac{\partial\mathcal F}{\partial\Omega}\frac{\partial\Omega}{\partial\phi_i}\right )^\top\frac{\partial\ell\left (\hat y^n, y^n\right )}{\partial\mathcal F}
  \end{aligned}
\end{equation}
where $\alpha$ is the learning rate, $N$ is batch size, and $t$ is the iteration. According to the chain rule used to find the gradient in backpropagation, the update of $\phi_i$ will influence the update of $\Omega$, and vice versa. According to Figure~\ref{inifig}(a) and \ref{inifig}(b), the gradient and the accuracy of the dominant modality will increase during the training process while the other two remain stable. Particularly, the gradient magnitude of the text modality increases very fast during the training process. This suggests the encoder of the dominant modality will be updated much faster than others, which makes $\frac{\partial \Omega}{\partial \phi}$ much larger. This phenomenon can also be validated by previous works~\citep{fan2023pmr, peng2022balanced, wu2022characterizing}. Besides, in Figure~\ref{inifig}(c), we present the gradient direction between each modality and the fusion. 
We can observe that the similarity between audio modality and the multimodal fusion is less than 0, indicating that they optimize towards the opposite direction, thus hindering the gradient update for multimodal branch. Meanwhile, the similarity between text modality and the multimodal fusion is increasing, suggesting the optimization direction towards the dominant modality.
With the progress of optimization, the encoder of the dominant modality can make relatively accurate predictions, which makes the fusion module $\Omega$ only depend on this modality (both magnitude and direction as mentioned above), leaving other encoders under-optimized.

\subsection{Classifier-guided Gradient Modulation}\label{s33}
\subsubsection{Gradient Magnitude Modulation}
As we discuss in Section~\ref{s32}, the gradient magnitude of the dominant modality increases fast during the training while the other modalities remain stable, thus being under-optimized. To balance the training process and make the fusion module benefit from all the encoders simultaneously, we propose the classifier-guided gradient modulation. Specifically, we use a modality-specific classifier to make predictions of $h_i$ in Equation~\ref{e1}. We can write the process as:
\begin{equation}\label{e5}
  \hat y_{m_i} = f_i(h_i)
\end{equation}
where $f_i$ is the classifier of modality $m_i$ and $\hat y_{m_i}$ is the prediction only using modality $m_i$. The classifier $f_i$ consists of 1-2 multi-head self-attention (MSA) layers~\citep{vaswani2017attention} and a fully connected layer for classification and regression tasks. And for segmentation tasks, $f_i$ is a light decoder. After $h_i$ is inputted into the fusion module $\Omega$, it becomes a more high-level representation. Therefore, we use several MSA layers to make $h_i$ more consistent with the output of the fusion module.

For a specific task, we have some evaluation metrics such as accuracy and mean absolute error. Here, we choose one of the evaluation metrics (\textit{e.g.} accuracy for classification tasks and mean absolute error for regression tasks) and denote it as $\varepsilon$. For every iteration of training, we can get predictions from the classifiers. We denote the predictions as $\boldsymbol{\hat y^i}=(\hat y_{m_1}^i,\hat y_{m_2}^i,\cdots,\hat y_{m_M}^i)$ where $i$ is the current iteration. Furthermore, we evaluate the task using $\boldsymbol{\hat y^i}$ to get the evaluation metric $\boldsymbol{\varepsilon^i}=(\varepsilon_{m_1}^i, \varepsilon_{m_2}^i, \cdots, \varepsilon_{m_M}^i)$. Here, we use the difference between the two consecutive $\boldsymbol{\varepsilon}$ to denote the modality-specific improvement for each iteration:
\begin{equation}\label{e6}
  \begin{aligned}
  \Delta {\boldsymbol{\varepsilon^{t+1}}} &= \boldsymbol{\varepsilon^{t+1}} - \boldsymbol{\varepsilon^t} = (\Delta \varepsilon^{t+1}_{m_1},\Delta \varepsilon^{t+1}_{m_2},\cdots,\Delta \varepsilon^{t+1}_{m_M})\\
  &= (\varepsilon_{m_1}^{t+1}-\varepsilon_{m_1}^t, \varepsilon_{m_2}^{t+1}-\varepsilon_{m_2}^t, \cdots, \varepsilon_{m_M}^{t+1}-\varepsilon_{m_M}^t)
  \end{aligned}
\end{equation}
where $t=0,1,2,\cdots,T$ and $T$ is the total iterations of training. Particularly, $\boldsymbol{\varepsilon^{0}}$ is initialized to $\boldsymbol{0}$. In some multimodal datasets, only using one of the modalities can achieve good results so we can not directly use $\boldsymbol{\varepsilon}$ to measure the utilization rate of different modalities. Therefore, it is reasonable to use the difference between $\boldsymbol{\varepsilon}$ to denote the relative improvements for each iteration. Then, we define the gradient magnitude balancing term of modality $m_i$ for the $t$-th iteration as follows:
\begin{equation}\label{e7}
  \mathcal B^{t}_{m_i} = \rho\frac{\sum_{k=1, k\neq i}^M\Delta \varepsilon^t_{m_k}}{\sum_{k=1}^M\Delta \varepsilon^t_{m_k}}
\end{equation}
where $\rho$ is a scaling hyperparameter and $M$ is the number of modalities. According to Equation~\ref{e7}, it is easy to find that when the performance of the model only using modality $m_i$ improves very fast (\textit{i.e.} $\Delta\varepsilon_{m_i}^t$ is large), $\mathcal B_{m_i}^t$ will be small. Similarly, when the modality $m_i$ brings relatively limited improvements to the model (\textit{i.e.} $\Delta\varepsilon_{m_i}^t$ is small), $\mathcal B_{m_i}^t$ will be large. Therefore, $\mathcal B_{m_i}^t$ is able to measure the relative utilization rate of these modalities and we can use $\mathcal B_{m_i}^t$ to modulate the magnitude of the gradient of the encoder $\phi_i$. So Equation~\ref{e4} can be rewritten as:
\begin{equation}\label{e8}
  \begin{aligned}
     \theta_{t+1}^{\phi_i}&=\theta_{t}^{\phi_i}-\alpha\mathcal B^{t+1}_{m_i}\nabla_{\theta^{\phi_i}}\mathcal L(\theta_{t}^{\phi_i}) \\
     &= \theta_{t}^{\phi_i}-\alpha\rho\frac{\sum_{k=1, k\neq i}^M\Delta \varepsilon^{t+1}_{m_k}}{\sum_{k=1}^M\Delta \varepsilon^{t+1}_{m_k}}\nabla_{\theta^{\phi_i}}\mathcal L(\theta_{t}^{\phi_i})
  \end{aligned}
\end{equation}
According to Equation~\ref{e2}, we know that the final predictions are closely related to $h_i$. Therefore, after we modulate the gradient of the corresponding encoder $\phi_i$, it has an impact on the input of $\Omega$, which in turn helps the optimization of the fusion module $\Omega$.

\begin{algorithm}[tb]
  \caption{Classifier-guided gradient modulation}
  \label{a1}
\begin{algorithmic}[1]
  \STATE {\bfseries Input:} Training dataset $\mathcal D=$ $\{(\boldsymbol{x_i}, y_i)\}_{i=1}^N$, iteration number $T$, the number of modalities $M$, model $F=(\phi_i,\Omega, \mathcal F)$, classifiers $f_i$, and hyperparameters.
  \STATE {\bfseries Initiate:} $\boldsymbol{\varepsilon^p}\leftarrow\boldsymbol{0}$, $\boldsymbol{\varepsilon^n}\leftarrow\text{Empty List}$, $\mathcal L_{gm}\leftarrow0$, classifier gradient list $L_g$.

  \FOR{$t=1$ {\bfseries to} $T$}
  \STATE $\mathcal D_t\stackrel{\text{Sample}}{\longleftarrow}\mathcal D$;
  \STATE Forward propagation to get representations $\boldsymbol{h}=(h_1,h_2,\cdots,h_M)$ in Equation~\ref{e1};
  \FOR{$i=1$ {\bfseries to} $M$}
  \STATE Make predictions with $h_i$ using Equation~\ref{e5};
  \STATE Calculate $\varepsilon^t_{m_i}$ and append it to $\boldsymbol{\varepsilon^n}$;
  \STATE Append $\nabla_{\theta^{f_i}}\mathcal L(\theta^{f_i})$ to $L_g$;
  \ENDFOR
  \STATE $\Delta \boldsymbol{\varepsilon^t}=\boldsymbol{\varepsilon^n}-\boldsymbol{\varepsilon^p}$;
  \STATE Calculate $\mathcal B_{m_i}^t,i=1,2,\cdots,M$ using Equation~\ref{e7};
  \STATE Calculate the loss $\mathcal L=\mathcal L_{task} + \lambda\mathcal L_{gm}$ and backward;
  \STATE Calculate $\mathcal L_{gm}^t$ using Equation~\ref{e12};
  \STATE $\boldsymbol{\varepsilon^p}\leftarrow\boldsymbol{\varepsilon^n}$, $\mathcal L_{gm}\leftarrow\mathcal L_{gm}^t$, $\boldsymbol{\varepsilon^n}\leftarrow\text{Empty List}$;
  \STATE Update parameters using Equation~\ref{e3} and \ref{e8}.
  \ENDFOR

\end{algorithmic}
\end{algorithm}

\subsubsection{Gradient Direction Modulation}\label{s332}
As \citet{wu2022characterizing} discover, when the model only depends on one modality to perform well, it does not continue to learn to use other modalities. As discussed in Section~\ref{s32}, it means that this modality dominates the updating of the model. Previous works~\citep{wu2022characterizing, peng2022balanced, li2023boosting} address this problem mainly by focusing on gradient magnitude modulation. However, in Section~\ref{s32}, we find that the model is optimized towards the dominant modality. Therefore, in this subsection, we introduce a method that could modulate the direction of the gradients to balance the training process.

In general, we want to balance the optimization direction of the model when the model only relies on one modality to make predictions. 
Therefore, we propose to enforce the gradient direction of the model as close as possible to the weighted average gradient direction of models only using one modality. We use $\mathcal B_{m_i}^t$ in Equation~\ref{e7} as the weight term. This ensures that when the model tends to optimize towards the dominant modality, $\mathcal B_{m_i}^t$ can help the model use information from other modalities. Besides, since $\mathcal B_{m_i}^t$ changes during the training process, this term can make a dynamic adjustment to balance the optimization directions. Concretely, we can feed one modality into the model and drop other modalities by replacing them with $\boldsymbol{0}$ or other fixed values during training to calculate the gradient of this modality. By this method, we can calculate the unimodal gradients for all modalities. Then, we just enforce the gradient direction of the model as close as possible to the weighted average of these unimodal gradient directions. However, this method is very complex during training, because in every iteration we need to drop modalities to calculate the unimodal gradients, which is time-consuming with the increase in the number of modalities.

Therefore, we propose to use the gradients of the classifiers $f_i$ to represent the unimodal gradients. We will later demonstrate they are similar (Section~\ref{s44} and Figure~\ref{ftsne}). Here, we take the gradient of regression tasks as an example where the output dimension is 1 so the gradient is an $n$-d vector. For classification tasks or other tasks where the gradient is a matrix, see Appendix~\ref{ap1} for details. Concretely, we can calculate the gradient of the classifier $f_i$ as:
\begin{equation}\label{e9}
  \begin{aligned}
     \nabla_{\theta^{f_i}}\mathcal L(\theta^{f_i}) &= \frac{\partial\mathcal L(\theta^{f_i})}{\partial f_i} = \left [\frac{\partial\mathcal L(\theta^{f_i})}{\partial\theta^{f_i}_1}, \frac{\partial\mathcal L(\theta^{f_i})}{\partial\theta^{f_i}_2}, \cdots, \frac{\partial\mathcal L(\theta^{f_i})}{\partial\theta^{f_i}_n} \right ]^\top
  \end{aligned}
\end{equation}
where $\theta^{f_i}$ represents the parameters of $f_i$. Different from $\theta^{f_i}_t$ in Equation~\ref{e3} and \ref{e4} where $t$ is the iteration, $\theta^{f_i}_n$ here represents one of the variables of $\theta^{f_i}$. Similarly, we can calculate the gradient of the classifier $\mathcal F$ of the fusion module as:
\begin{equation}\label{e10}
  \begin{aligned}
     \nabla_{\theta^{\mathcal F}}\mathcal L(\theta^{\mathcal F}) &= \frac{\partial\mathcal L(\theta^{\mathcal F})}{\partial\mathcal F} = \left [\frac{\partial\mathcal L(\theta^{\mathcal F})}{\partial\theta^{\mathcal F}_1}, \frac{\partial\mathcal L(\theta^{\mathcal F})}{\partial\theta^{\mathcal F}_2}, \cdots, \frac{\partial\mathcal L(\theta^{\mathcal F})}{\partial\theta^{\mathcal F}_n} \right ]^\top
  \end{aligned}
\end{equation}
We regard $\nabla_{\theta^{f_i}}\mathcal L,i=1,2,\cdots,M$ as the unimodal gradients and $\nabla_{\theta^{\mathcal F}}\mathcal L$ as the fusion gradients. As mentioned before, we want to make $\nabla_{\theta^{\mathcal F}} \mathcal L$ as close as possible to the weighted average direction of $\nabla_{\theta^{f_i}}\mathcal L,i=1,2,\cdots,M$. Let $\textrm{sim}(\boldsymbol{u},\boldsymbol{v})=\boldsymbol{u}^\top\boldsymbol{v}/||\boldsymbol{u}||||\boldsymbol{v}||$ denote the dot product between $\ell_2$ normalized $\boldsymbol{u}$ and $\boldsymbol{v}$ (\textit{i.e.} cosine similarity). We can enforce the gradient direction of the fusion module as close as possible to the weighted average of these unimodal gradient directions by maximizing their cosine similarity:
\begin{equation}\label{e11}
  \max \sum_{i=1}^M\mathcal B_{m_i}^t\textrm{sim}\left (\nabla_{\theta^{\mathcal F}}\mathcal L, \nabla_{\theta^{f_i}}\mathcal L\right )
\end{equation}
where $t$ is the current iteration. We rewrite Equation~\ref{e11} as a loss term:
\begin{equation}\label{e12}
  \mathcal L_{gm}^t = \frac{1}{M}\sum_{i=1}^M|\mathcal B_{m_i}^t|-\mathcal B_{m_i}^t\textrm{sim}\left (\nabla_{\theta^{\mathcal F}_t}\mathcal L, \nabla_{\theta^{f_i}_t}\mathcal L\right )
\end{equation}
The cosine similarity is a number between $-1$ and $1$. By adding $|\mathcal B_{m_i}^t|$ to the loss term, we can ensure that the loss $\mathcal L_{gm}$ is always positive. As aforementioned, when modality $m_i$ has limited improvement, $\mathcal B_{m_i}^t$ is large. Therefore, the corresponding term in $\mathcal L_{gm}^t$ will be large, making the optimization direction towards modality $m_i$, which will balance the learning process.

Then our overall loss function can be written as:
\begin{equation}\label{e13}
  \mathcal L^t=\mathcal L_{task}+\lambda\mathcal L_{gm}^t
\end{equation}
where $\mathcal L_{task}$ is the task loss function (\textit{e.g.} cross-entropy loss, L1 loss) and $\lambda$ is a trade-off between the two loss terms. We present our overall method in Algorithm~\ref{a1}.

\begin{wraptable}{r}{7.5cm}
  \vskip -0.8in
  \caption{The difference between the four datasets we use.}
  \label{t1}
  \begin{center}
     \begin{small}
          \begin{tabular}{ccc}
        \toprule
        Dataset&Task type&No. of modality\\
        \midrule
        UPMC-Food 101&Classification&2\\
        CMU-MOSI&Regression&3\\
        IEMOCAP&Classification&3\\
        BraTS 2021&Segmentation&4\\
        \bottomrule
     \end{tabular}
  \end{small}
  \end{center}
  \vskip -0.2in
\end{wraptable}

\section{Experiments}
\subsection{Datasets and Evaluation Metrics}
We use four multimodal datasets: UPMC-Food 101~\citep{7169757}, CMU-MOSI~\citep{mosi}, IEMOCAP~\citep{Busso2008IEMOCAPIE}, and BraTS 2021~\citep{baid2021rsna}. Table~\ref{t1} presents the difference between these four datasets.
 
\noindent\textbf{UPMC-Food 101}~\citep{7169757} is a food classification dataset, which contains about 100,000 recipes for a total of 101 food categories. Each item in the dataset is represented by one image plus textual information. 
We use accuracy and F1 score to evaluate the performance of the model.

\noindent\textbf{CMU-MOSI}~\citep{mosi} is a popular dataset for multimodal (audio, text and video) sentiment analysis.  
Each video segment is manually annotated with a sentiment score ranging from strongly negative to strongly positive (-3 to +3). Following previous work~\citep{tsai-etal-2019-multimodal,guo2024multimodal}, we use binary accuracy (ACC-2), F1 score, 7-class accuracy (ACC-7), mean absolute error (MAE) and pearson correlation (Corr) to evaluate the performance of the model.

\begin{minipage}{\textwidth}
  \centering
 \begin{minipage}[t]{0.4\textwidth}
  \centering
     \makeatletter\def\@captype{table}\makeatother\caption{Quantitative results on the UPMC-Food 101 dataset. \textbf{Bold}: best results. \underline{Underline}: second best results.}\label{rfood}
     \vskip 0.2in
     \resizebox{0.9\columnwidth}{!}{%
     \begin{tabular}{lcc}
      \toprule
      Method & Acc & F1 \\
      \midrule
      Text only    & 84.77& 84.72 \\
      Image only & 68.24& 68.23 \\
      \midrule
      Baseline    & 90.32& 90.30 \\
      G-Blending~\citep{wang2020makes}   & 90.42& 90.38\\
      Greedy~\citep{wu2022characterizing}  & 91.21 & 91.20\\
      OGE~\citep{peng2022balanced}     & 91.08& 91.08\\
      AGM~\citep{li2023boosting}     & 91.49& 91.48\\
      PMR~\citep{fan2023pmr} & 92.01& 91.98\\
      UMT~\citep{du2023uni}&91.82&91.82\\
      UME~\citep{du2023uni}&90.69&90.68\\
      QMF~\citep{zhang2023provable}&\underline{92.87}&\underline{92.85}\\
      ReconBoost~\citep{hua2024reconboost}&92.52&92.51\\
      \midrule
      CGGM      &\bf{92.94} & \bf{92.90} \\
      \bottomrule
      \end{tabular}}
  \end{minipage}
  \hskip 0.2in
  \begin{minipage}[t]{0.55\textwidth}
   \centering
        \makeatletter\def\@captype{table}\makeatother\caption{Results on BraTS 2021. WT, TC and ET denote the dice score of Whole Tumor, Tumor Core and Enhancing Tumor respectively.}\label{rbrats}
        \vskip 0.25in
        \begin{tabular}{lcccc}
          \toprule
          Method & WT & TC & ET & Avg. \\
          \midrule
          flair only   &70.42&51.41&45.27&55.70 \\
          t1 only &50.73&36.13&38.77&41.88\\
          t2 only   &64.82&42.17&34.62&47.20 \\
          t1ce only  &56.61&53.83&53.14&54.53\\
          \midrule
          Baseline    &74.03&67.08&66.53&69.21\\
          MRD &74.02&68.04&67.28&69.78 \\
          MSLR &74.47&69.98&67.20&70.55 \\
          UMT~\citep{du2023uni}&74.15&67.69&66.80&69.55\\
          UME~\citep{du2023uni}&74.65&68.74&67.58&70.32\\
          QMF~\citep{zhang2023provable}&75.11&\underline{70.78}&68.94&71.61\\
          ReconBoost~\citep{hua2024reconboost}&\underline{75.21}&70.18&\underline{70.01}&\underline{71.80}\\
          \midrule
          CGGM      &\bf{76.94}&\bf{72.75}&\bf{72.14}&\bf{73.94}\\
          \bottomrule
        \end{tabular}
   \end{minipage}
\end{minipage}
\vskip 0.2in

\begin{table*}
  \caption{Quantitative results on the CMU-MOSI and IEMOCAP datasets. \textbf{Bold}: best results. \underline{Underline}: second best results.}
  \label{rmsa}
  \vskip 0.15in
  \begin{center}
  \begin{small}
  \begin{tabular}{lccccccc}\toprule
  \multirow{2}{*}{Method}&\multicolumn{5}{c}{\textbf{CMU-MOSI}}&\multicolumn{2}{c}{\textbf{IEMOCAP}}\\
  \cmidrule(lr){2-6}\cmidrule(lr){7-8}
  &Acc-2&Acc-7&F1&MAE&Corr&Acc&F1\\
  \midrule
  Text only&76.83&28.24&76.80&1.016&0.664&65.35&64.30\\
  Audio only&64.12&23.04&66.96&1.451&0.510&52.18&50.14\\
  Video only&62.00&21.65&66.18&1.441&0.499&54.55&53.97\\
  \midrule
  Baseline&81.23&29.26&81.23&0.952&0.710&70.74&69.53\\
  MRD&80.78&31.44&80.75&0.975&0.688&71.81&71.08\\
  MSLR&81.22&31.00&81.19&0.950&0.704&71.99&70.96\\
  AGM~\citep{li2023boosting}&-&-&-&-&-&72.35&71.94\\
  UMT~\citep{du2023uni}&\underline{81.78}&\underline{31.98}&\underline{81.77}&\underline{0.942}&\underline{0.712}&70.75&69.81\\
  UME~\citep{du2023uni}&80.83&30.94&80.78&0.969&0.701&71.53&70.94\\
  QMF~\citep{zhang2023provable}&-&-&-&-&-&72.08&71.64\\
  ReconBoost~\citep{hua2024reconboost}&-&-&-&-&-&\underline{73.14}&\underline{72.71}\\
  \midrule
  CGGM&\bf{82.84}&\bf{33.73}&\bf{82.74}&\bf{0.915}&\bf{0.717}&\bf{75.38}&\bf{74.97}\\
  \bottomrule
  \end{tabular}
  \end{small}
  \end{center}
  \vskip -0.3in
\end{table*}

\noindent\textbf{IEMOCAP}~\citep{Busso2008IEMOCAPIE} is a multimodal emotion recognition dataset, which contains recorded videos from ten actors in five dyadic conversation sessions. 
Following previous works~\citep{tsai-etal-2019-multimodal,wang2018words,guo2024multimodal}, four emotions (happiness, anger, sadness and neutral state) are selected for emotion recognition. We use accuracy and F1 score to evaluate the performance of the model.

\noindent\textbf{BraTS 2021}~\citep{baid2021rsna} is a 3D multimodal brain tumor segmentation dataset, which has four modalities: \textit{flair, t1ce, t1} and \textit{t2}. The input image size is $240\times240\times155$. The annotations are combined into three nested subregions: Whole Tumor (WT), Tumor Core (TC), and Enhancing Tumor (ET). We use Dice score of these three nested subregions and their average value to evaluate the performance.

\subsection{Implementation Details}
\noindent\textbf{Input.} For UPMC-Food 101, we use extracted features as inputs. Specifically, we use the pre-trained bert-base-uncased model~\citep{devlin2018bert} to extract text features and use pre-trained ViT~\citep{dosovitskiy2020image} on ImageNet to extract image features. For CMU-MOSI and IEMOCAP, we follow \citet{guo2024multimodal} to extract acoustic, visual and textual features. For BraTS 2021, we use the preprocessed raw images as inputs.

\noindent\textbf{Backbone.} For UPMC-Food 101, CMU-MOSI and IEMOCAP, we use transformer encoders~\citep{vaswani2017attention} as modality encoders and the fusion module. For the BraTS 2021 dataset, we use DeepLab v3+~\citep{chen2018encoder} as the encoders and several convolution layers as the fusion module.

\noindent\textbf{Training Details.} For images in UPMC-Food 101 and BraTS 2021, we implement data augmentation strategies, including random cropping, random flipping, color jitter, adding noise, etc. To save memory, we consider BraTS 2021 as a 2D segmentation task by randomly slicing an image from the 3D image. For CMU-MOSI, we use L1 loss as our loss function. For UPMC-Food 101 and IEMOCAP, we use cross-entropy loss. For BraTS 2021, we use the combination of soft dice loss and cross-entropy loss. Besides, we use the Adam optimizer for CMU-MOSI, the AdamW optimizer for UPMC-Food 101 and IEMOCAP, and the SGD optimizer for BraTS 2021. Other hyperparameters are described in Appendix~\ref{ap2} in detail.

\begin{figure}
  \vskip 0.2in
  \begin{center}
  \subfigure[]{\includegraphics[width=.32\columnwidth]{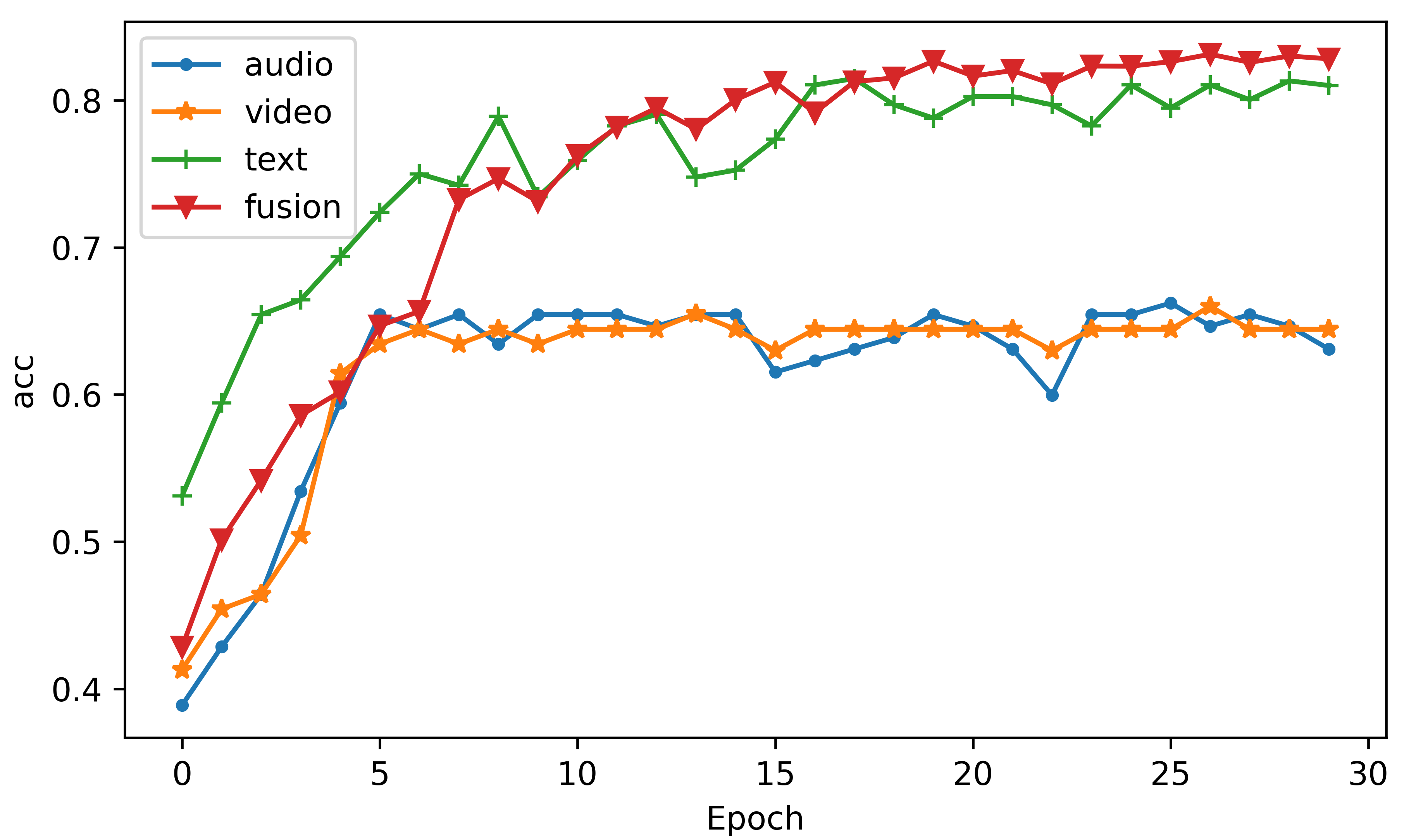}}\hspace{5pt}
 \subfigure[]{\includegraphics[width=.32\columnwidth]{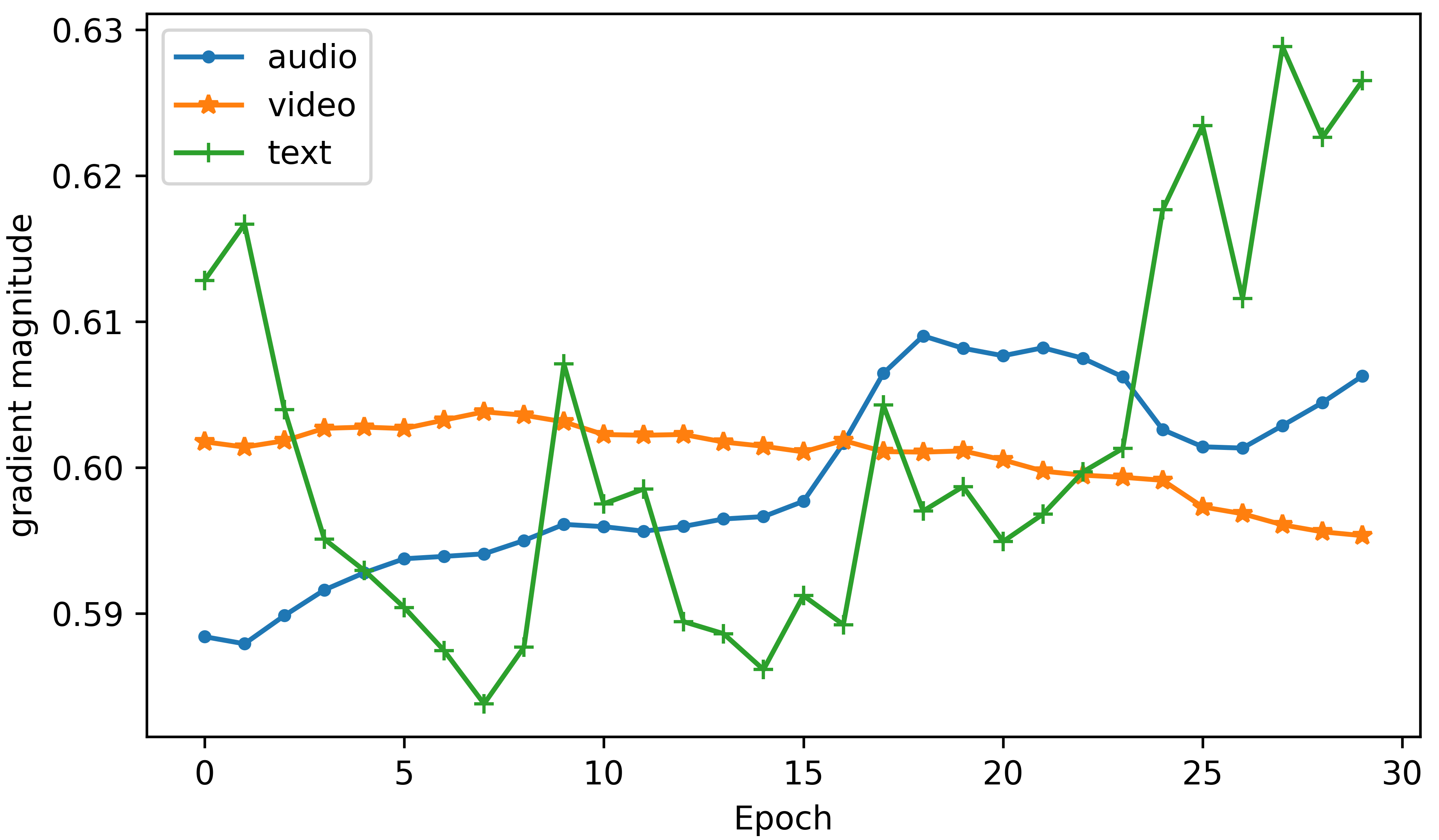}}\hspace{5pt}
 \subfigure[]{\includegraphics[width=.32\columnwidth]{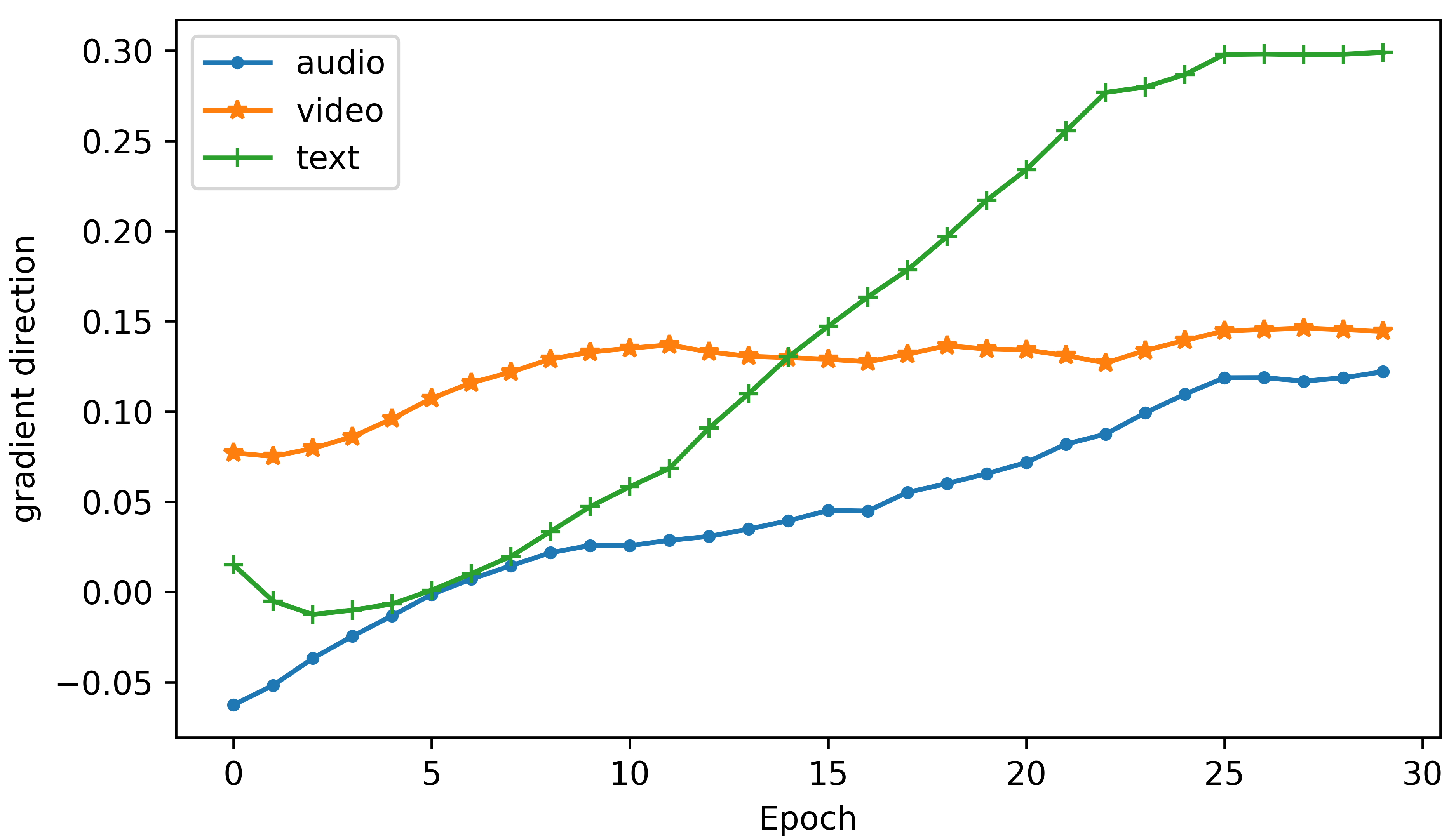}}\hspace{5pt}
\\
  \caption{Changes in (a) performance, (b) gradient magnitude and (c) direction during training with CGGM. We get the results on CMU-MOSI dataset.}
  \label{change}
  \end{center}
  \vskip -0.2in
\end{figure}

\begin{figure}
  \vskip 0.2in
  \begin{center}
  \subfigure[UPMC-Food 101]{\includegraphics[width=.23\columnwidth]{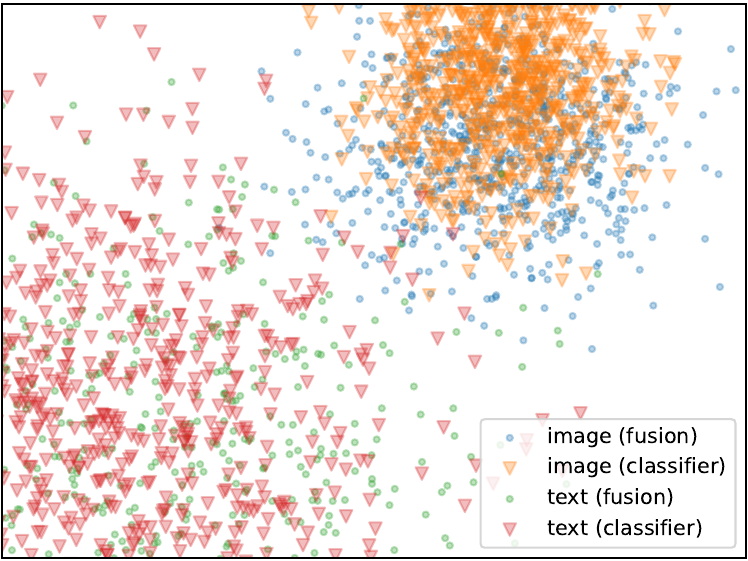}}\hspace{5pt}
 \subfigure[CMU-MOSI]{\includegraphics[width=.23\columnwidth]{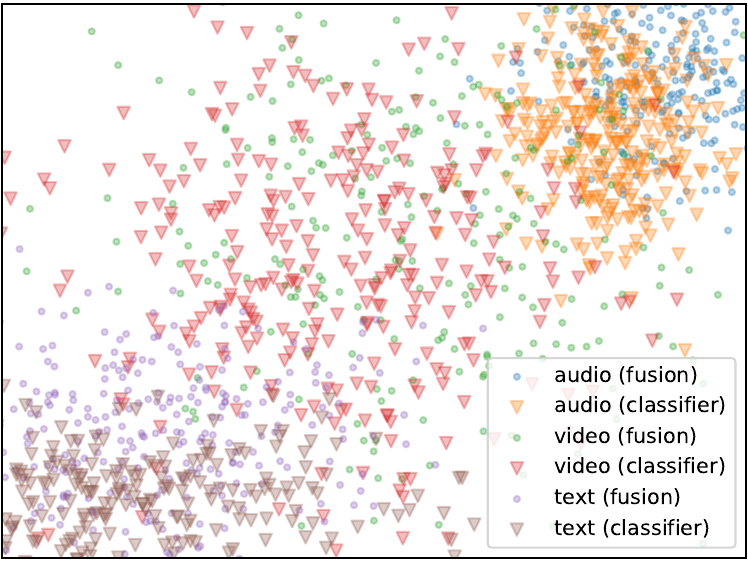}}
 \subfigure[IEMOCAP]{\includegraphics[width=.23\columnwidth]{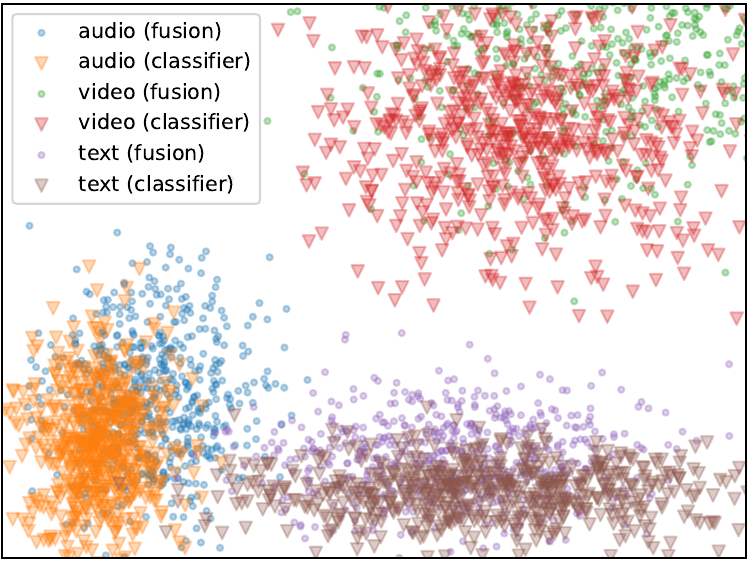}}\hspace{5pt}
 \subfigure[BraTS 2021]{\includegraphics[width=.23\columnwidth]{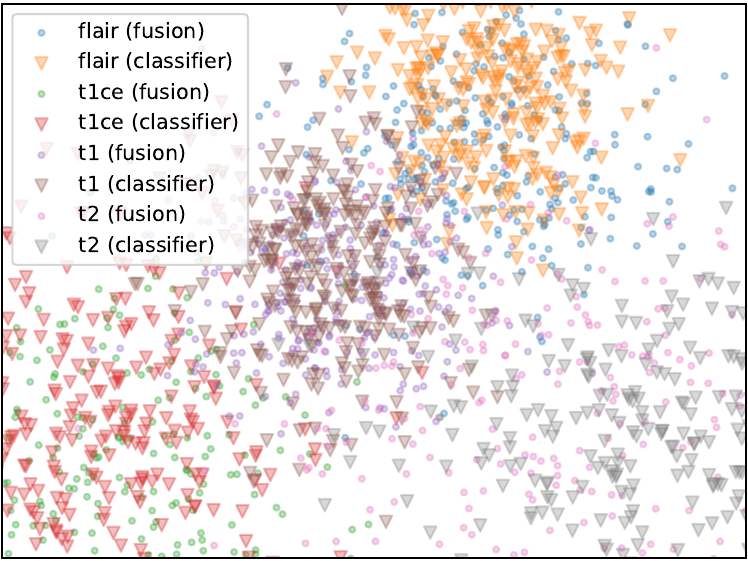}}\\
  \caption{t-SNE visualization of the gradients of classifiers and the unimodal gradients. Each point represents a gradient vector or matrix of a batch of data.}
  \label{ftsne}
  \end{center}
  \vskip -0.1in
\end{figure}

\subsection{Main Results}
\textbf{Comparison with the state-of-the-arts.} We compare our CGGM with other methods to demonstrate the effectiveness of our proposed method. For these four datasets, we compare CGGM with the model training only using one modality, multimodal joint training (Baseline), Modality Random Dropout (MRD), and Modality-specific Learning Rate (MSLR) methods. Additionally, we compare CGGM with SOTA methods including G-Blending~\citep{wang2020makes}, Greedy~\citep{wu2022characterizing}, OGM~\citep{peng2022balanced}, AGM~\citep{li2023boosting}, PMR~\citep{fan2023pmr}, UMT~\citep{du2023uni}, UME~\citep{du2023uni}, QMF~\citep{zhang2023provable} and ReconBoost~\citep{hua2024reconboost}.
Table~\ref{rfood}, \ref{rbrats} and \ref{rmsa} present the results of CGGM and its compared methods. Compared with the baseline method, CGGM brings significant improvements to the performance of the model, which demonstrates the effectiveness of our proposed method. Besides, CGGM takes both gradient magnitude and direction into consideration, thus making it outperform other gradient modulation methods consistently in all four datasets. Most importantly, CGGM can be easily applied to various tasks and has good performance, including classification tasks, regression tasks, segmentation tasks, etc. Meanwhile, CGGM has no limitations on the optimizer, loss function and the number of modalities.

\textbf{Effectiveness of CGGM.} In Figure~\ref{change}, we visualize the changes in accuracy, gradient magnitude and direction during training with CGGM. Compared with Figure~\ref{inifig}(a), the accuracy of text modality in Figure~\ref{change}(a) does not increase very fast with CGGM, which indicates that CGGM imposes constraints to the dominant modality during the optimization process. Besides, the accuracies of all the modalities and the fusion improves, indicating the effectiveness of CGGM. Additionally, in Figure~\ref{inifig}(b), the dominant modality always has the largest gradient while in Figure~\ref{change}(b), the gradient magnitude of the text modality decreases at first, indicating that CGGM slows down its optimization and accelerates other modalities' optimization, helping each modality learn sufficiently, thus improving the multimodal performance. In case of gradient direction, in Figure~\ref{inifig}, the similarity between audio modality and the fusion is always less than 0 during the training process, indicating an opposite optimization direction between the unimodal and multimodal, thus hindering the optimization process. In Figure~\ref{change}, we observe the multimodal direction is consistent with all modalities, indicating that the multimodal branch utilizes unimodal information efficiently.

\begin{table}[]
  \centering
  \caption{Accuracy on IEMOCAP. $f_1,f_2$ and $f_3$ represent the audio, video and text classifier, respectively. We train three separate models in unimodal training.}
  \label{cls_acc}
  \begin{tabular}{@{}ccc|ccc|ccc@{}}
  \toprule
  \multicolumn{3}{c}{unimodal training} & \multicolumn{3}{c}{multimodal training} & \multicolumn{3}{c}{CGGM} \\ \cmidrule(lr){1-3}\cmidrule(lr){4-6}\cmidrule(lr){7-9}
  audio & video & text & $f_1$ & $f_2$ & $f_3$ & $f_1$ & $f_2$ & $f_3$ \\\midrule
  52.18 & 54.55 & 65.35 & 50.59 & 53.04 & 67.15 & 54.95 & 56.77 & 67.39 \\ \bottomrule
  \end{tabular}%
  \vskip -0.1 in
\end{table}

\subsection{Classifier Performance and Gradient Direction}\label{s44}
\noindent\textbf{Classifier performance.} In Table~\ref{cls_acc}, we present the accuracy of classifiers in different situations. Unimodal training can be considered a baseline that fully utilizes the unimodal information. Compared with unimodal training, the accuracies of $f_1$ and $f_2$ in multimodal training drop slightly while the accuracy of $f_3$ increases slightly. This demonstrates that multimodal training can not fully utilize the information from audio and video, indicating that they are under-optimized. The improvement of $f_3$ indicates that the text modality is fully exploited and learns some information from the other two modalities. In contrast, the accuracies of the three classifiers in CGGM all improve. This suggests that during the balancing process, the fusion can fully utilize the information from all the modalities, thus in turn making the encoders of three modalities fuse the information from other modalities during backpropagation. Therefore, the accuracies of all the three classifiers improve correspondingly. This also validates the effectiveness of CGGM.

\noindent\textbf{Classifier gradient direction.} In Section~\ref{s332}, we propose to use the gradients of classifiers to represent the unimodal gradients. In this subsection, we give a visualization of the gradients of classifiers and the unimodal gradients. Specifically, for every batch of data, we input them into the model to get representations $h_i$ which are then fed into the classifiers $f_i$ to get the gradients of classifiers. Then we input $h_i$ of only one modality into the fusion module $\Omega$ to get the unimodal gradients. We use t-SNE~\citep{vandermaaten08a} to visualize the gradient vectors. Figure~\ref{ftsne} shows the visualization results on the four datasets. As shown in the figure, for each modality, the unimodal gradient vectors and the gradient vectors of the corresponding classifier are very close to each other, demonstrating that it is meaningful to use the gradients of the classifiers to represent the unimodal gradients.

\begin{figure*}
  \begin{center}
  \centerline{\includegraphics[width=0.87\columnwidth]{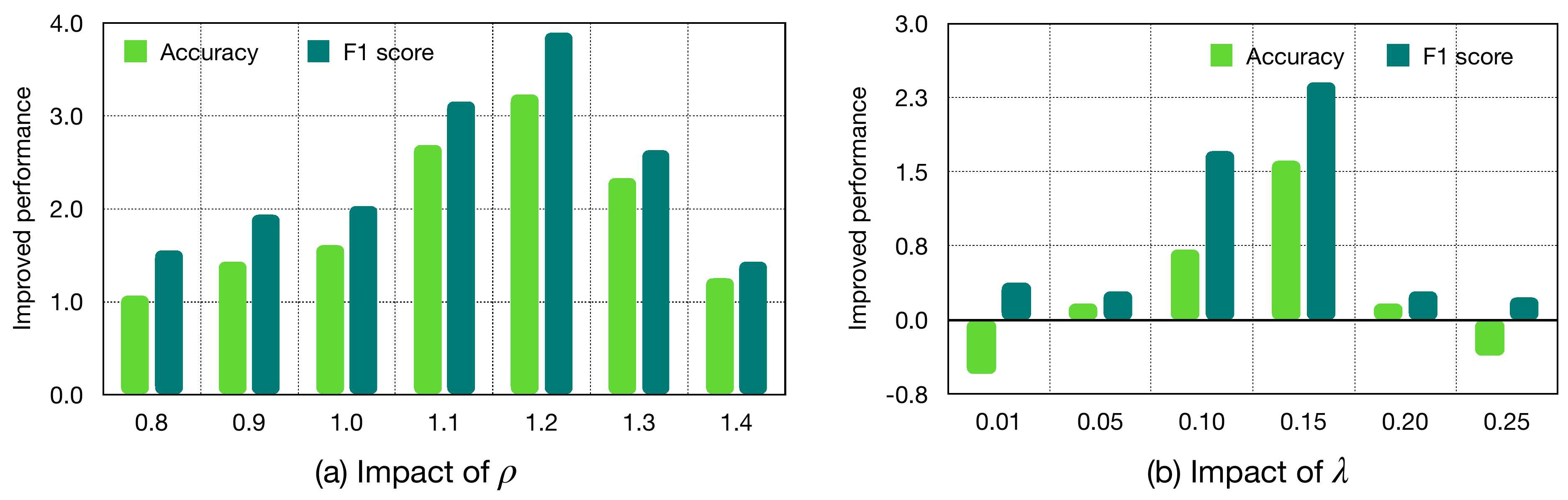}}
  \caption{The improved performance with different $\rho$ and $\lambda$ compared to the joint training baseline.}
  \label{abex}
  \end{center}
  \vskip -0.3in
\end{figure*}

\subsection{Ablation Study}
\noindent\textbf{Gradient magnitude and direction.} To measure the contribution of gradient magnitude modulation and gradient direction modulation separately, we present our ablation results on IEMOCAP in Table~\ref{ab:gmd}. Compared with the baseline in the first row, modulating the magnitude of the gradients brings more improvements to the performance of the model than modulating the direction of the gradients. Compared with the CGGM performance in Table~\ref{rmsa}, the combination of modulating gradient magnitude and gradient directions furthermore enhances the performance of the model. 

\begin{wraptable}{r}{7.5cm}
  \vskip -0.1in
  \caption{The benefits of modulating the magnitude of the gradients and the directions of the gradients.}
  \label{ab:gmd}
  \begin{center}
     \begin{small}
      \begin{tabular}{cccc}
        \toprule
        Model & Acc & F1 \\
        \midrule
        Baseline& 70.74& 69.53 \\
        CGGM ($\rho=1.0,\lambda=0$)& 72.35& 71.56 \\
        CGGM ($\rho=\text{None},\lambda=0.1$) & 72.41& 72.07 \\
        CGGM ($\rho=1.0,\lambda=0.1$) & 73.74 & 73.18 \\
        \bottomrule
      \end{tabular}
  \end{small}
  \end{center}
  \vskip -0.2in
\end{wraptable}

\noindent\textbf{Scaling hyperparameter $\rho$.} To explore the impacts of the scaling hyperparameter $\rho$ on the model's performance, we select seven different values of $\rho$ and present our results on IEMOCAP in Figure~\ref{abex}(a). We discover that the accuracy improves with the increase of $\rho$ before hitting the highest value when $\rho=1.2$. Then, the accuracy drops with the increase of $\rho$. Compared to the baseline, modulating the magnitude of the gradients brings consistent improvements regardless of how big $\rho$ is taken. Intuitively, the larger the $\rho$, the larger the modification to the gradients. Therefore, the results in the table indicate that we need to carefully choose a $\rho$ to avoid modifications that are too large or too small.

\noindent\textbf{Loss trade-off $\lambda$.} $\lambda$ measures the strength we modulate the directions of the gradients. We select six different values of $\lambda$ and present the results on IEMOCAP in Figure~\ref{abex}(b). As shown in the figure, when $\lambda$ is 0.01 or 0.25, the accuracy will decrease slightly. When $\lambda$ is too small, the modulation is insufficient and could influence the optimization process. When $\lambda$ is too large, the modulation is large and will influence the task loss, thus making optimization deviate from the task.

\section{Conclusion}
In this paper, we propose CGGM, a novel strategy to balance the multimodal training process. Compared to existing gradient modulation methods, CGGM has no limitations on the loss functions, the optimizer, the number of modalities, etc. Moreover, we consider both the magnitude and direction of the gradients with the guidance of the classifiers. Extensive experiments and ablation studies fully demonstrate the effectiveness and universality of CGGM. However, CGGM also has a limitation. CGGM needs to leverage extra classifiers to implement gradient modulation. Although these classifiers are lightweight, they still lead to more computational resources. We lead this challenging problem to future work.

\section*{Acknowledgement}
This work was supported by National Key R\&D Program of China under Grant No.2022ZD0162000.

\bibliography{example_paper}
\bibliographystyle{plainnat}

\newpage
\appendix

\section{Gradient Direction Modulation Details}\label{ap1}
For classification tasks, the classifier $f_i$ outputs more than one value. For example, for the UPMC-Food 101 dataset, $f_i$ output 101 values for each piece of data. Therefore, we can define $f_i$ as $f_i=(f_i^{(1)}, f_i^{(2)}, \cdots, f_i^{(m)})$ where $m$ is the number of output. We calculate the gradients of the classifiers $f_i$ as:
\begin{equation}
   \begin{aligned}
      \nabla_{\theta^{f_i}}\mathcal L(\theta^{f_i}) &= \frac{\partial\mathcal L(\theta^{f_i})}{\partial f_i}
      = 
       \left[
         \begin{array}{cccc}
            \frac{\partial\mathcal L(\theta^{f_i^{(1)}})}{\partial\theta^{f_i}_1} & \frac{\partial\mathcal L(\theta^{f_i^{(2)}})}{\partial\theta^{f_i}_1} & \cdots & \frac{\partial\mathcal L(\theta^{f_i^{(m)}})}{\partial\theta^{f_i}_1} \\
            \frac{\partial\mathcal L(\theta^{f_i^{(1)}})}{\partial\theta^{f_i}_2} & \frac{\partial\mathcal L(\theta^{f_i^{(2)}})}{\partial\theta^{f_i}_2} & \cdots & \frac{\partial\mathcal L(\theta^{f_i^{(m)}})}{\partial\theta^{f_i}_2} \\
            \vdots & \vdots & \cdots & \vdots \\
            \frac{\partial\mathcal L(\theta^{f_i^{(1)}})}{\partial\theta^{f_i}_n} & \frac{\partial\mathcal L(\theta^{f_i^{(2)}})}{\partial\theta^{f_i}_n} & \cdots & \frac{\partial\mathcal L(\theta^{f_i^{(m)}})}{\partial\theta^{f_i}_n} 
         \end{array}
         \right]
   \end{aligned}
\end{equation}
Similarly, the gradients of the fusion module classifier can be calculated as:
\begin{equation}
   \begin{aligned}
      \nabla_{\theta^{\mathcal F}}\mathcal L(\theta^{\mathcal F}) &= \frac{\partial\mathcal L(\theta^{\mathcal F})}{\partial \mathcal F}
      = 
       \left[
         \begin{array}{cccc}
            \frac{\partial\mathcal L(\theta^{\mathcal F^{(1)}})}{\partial\theta^{\mathcal F}_1} & \frac{\partial\mathcal L(\theta^{\mathcal F^{(2)}})}{\partial\theta^{\mathcal F}_1} & \cdots & \frac{\partial\mathcal L(\theta^{\mathcal F^{(m)}})}{\partial\theta^{\mathcal F}_1} \\
            \frac{\partial\mathcal L(\theta^{\mathcal F^{(1)}})}{\partial\theta^{\mathcal F}_2} & \frac{\partial\mathcal L(\theta^{\mathcal F^{(2)}})}{\partial\theta^{\mathcal F}_2} & \cdots & \frac{\partial\mathcal L(\theta^{\mathcal F^{(m)}})}{\partial\theta^{\mathcal F}_2} \\
            \vdots & \vdots & \cdots & \vdots \\
            \frac{\partial\mathcal L(\theta^{\mathcal F^{(1)}})}{\partial\theta^{\mathcal F}_n} & \frac{\partial\mathcal L(\theta^{\mathcal F^{(2)}})}{\partial\theta^{\mathcal F}_n} & \cdots & \frac{\partial\mathcal L(\theta^{\mathcal F^{(m)}})}{\partial\theta^{\mathcal F}_n} 
         \end{array}
         \right]
   \end{aligned}
\end{equation}
In order to calculate cosine similarity between these two terms, we flatten them into vectors and then use Equation~\ref{e12} to calculate $\mathcal L_{gm}$.

\section{Implementation Details}\label{ap2}

\begin{table*}[ht]
   \caption{Main hyperparameters of the four datasets.}
   \label{hyp}
   \centering
   \vskip 0.15in
   \begin{tabular}{c|c|c|c|c}\toprule
      Hyperparameters&UMPC-Food 101&CMU-MOSI&IEMOCAP&BraTS 2021\\
      \midrule
      batch size&128&64&32&64\\
      optimizer&AdamW&Adam&AdamW&SGD\\
      base lr&1e-3&1e-3&1e-3&1e-2\\
      classifier lr&5e-4&5e-4&5e-4&6e-3\\
      weight decay&3e-3&-&-&3e-4\\
      gradient clip&0.8&0.8&0.8&-\\
      scheduler&StepLR&StepLR&StepLR&CosineLR\\
      $\rho$&1.3&1.3&1.3&1.2\\
      $\lambda$&0.15&0.20&0.15&0.10\\
      warm-up epoch&-&-&-&10\\
      epoch&80&30&30&150\\
      \bottomrule
   \end{tabular}
   \vskip 0.2in
\end{table*}

Table~\ref{hyp} presents the main hyperparameters of the four datasets. Apart from the hyperparameters in the table, there are some task-specific hyperparameters.

For BraTS 2021, the start learning rate is set to 4e-4 with warm-up epochs to 1e-2 and the final learning rate is 1e-3. Besides, for the loss function, we use the combination of soft dice loss and cross-entropy loss, which can be represented as $\mathcal L_{task}=\mathcal L_{Dice}+\lambda_1 \mathcal L_{CE}$. We set $\lambda_1$ to 1. Particularly, we use a weighted cross-entropy loss function, where the weight is 0.2, 0.3, 0.25 and 0.25 for the background, label 1, label 2 and label 3, respectively.

\begin{figure}
  \begin{center}
  \subfigure[w/o. CGGM]{\includegraphics[width=.43\columnwidth]{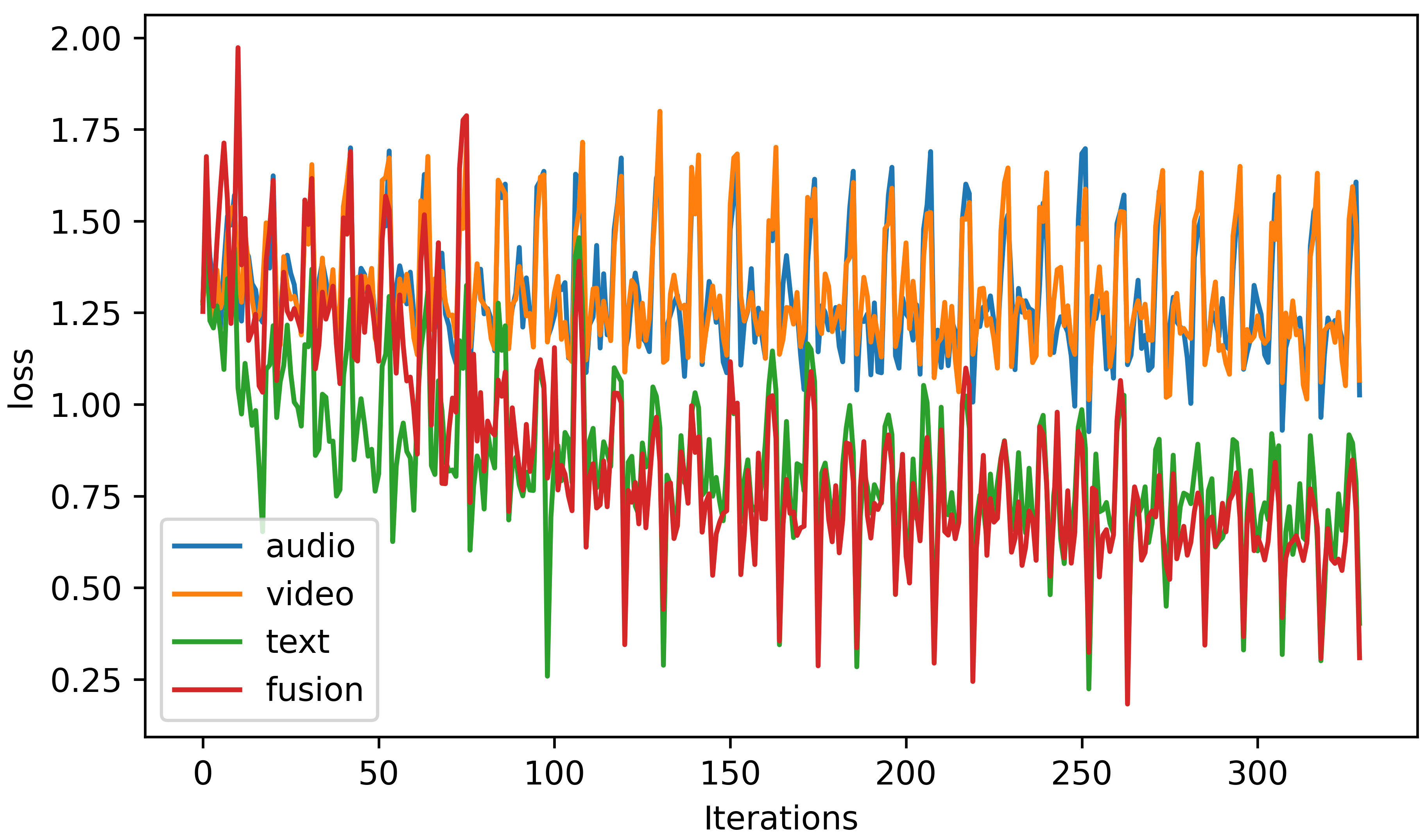}}\hspace{5pt}
 \subfigure[gradient magnitude]{\includegraphics[width=.43\columnwidth]{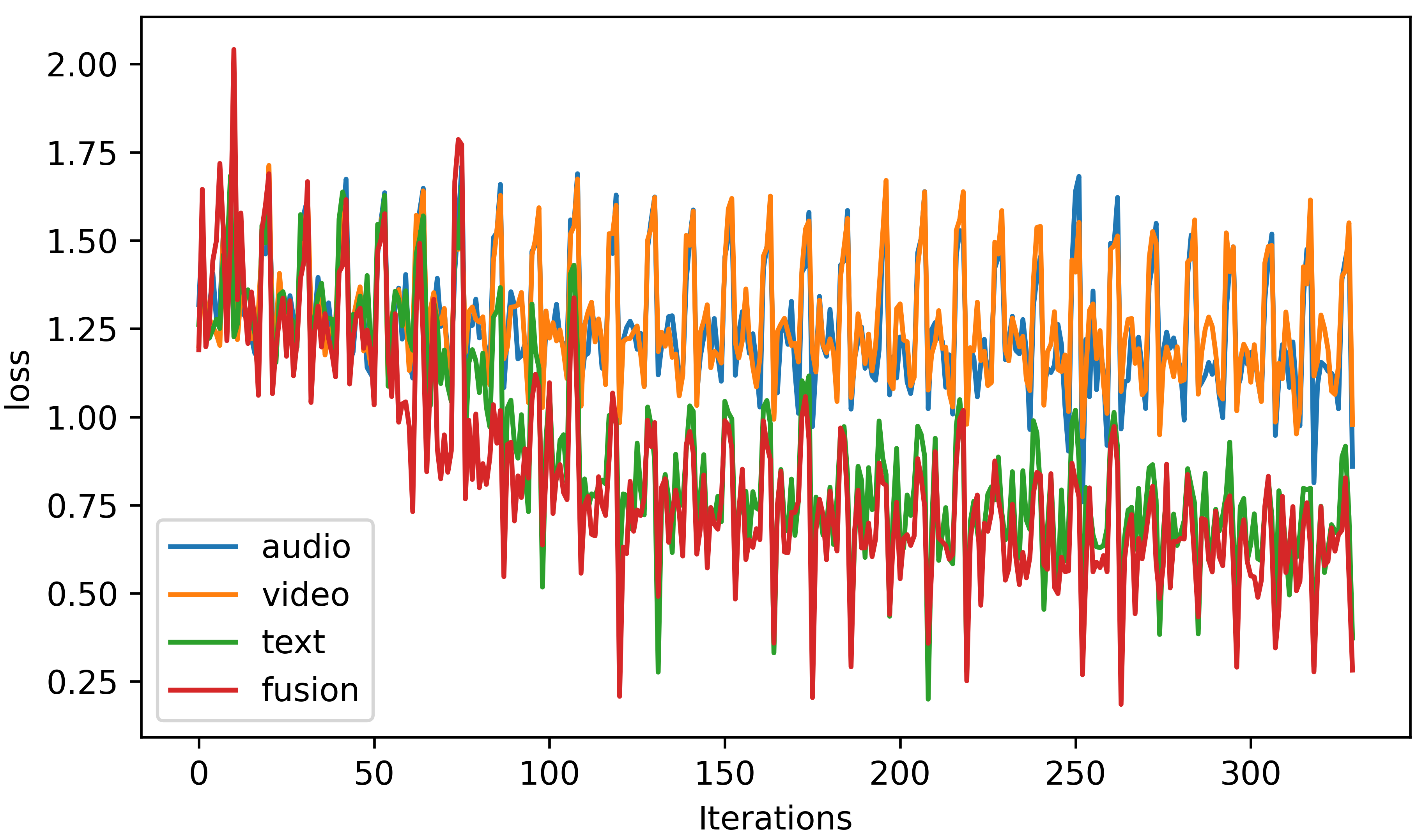}}\\
 \subfigure[gradient direction]{\includegraphics[width=.43\columnwidth]{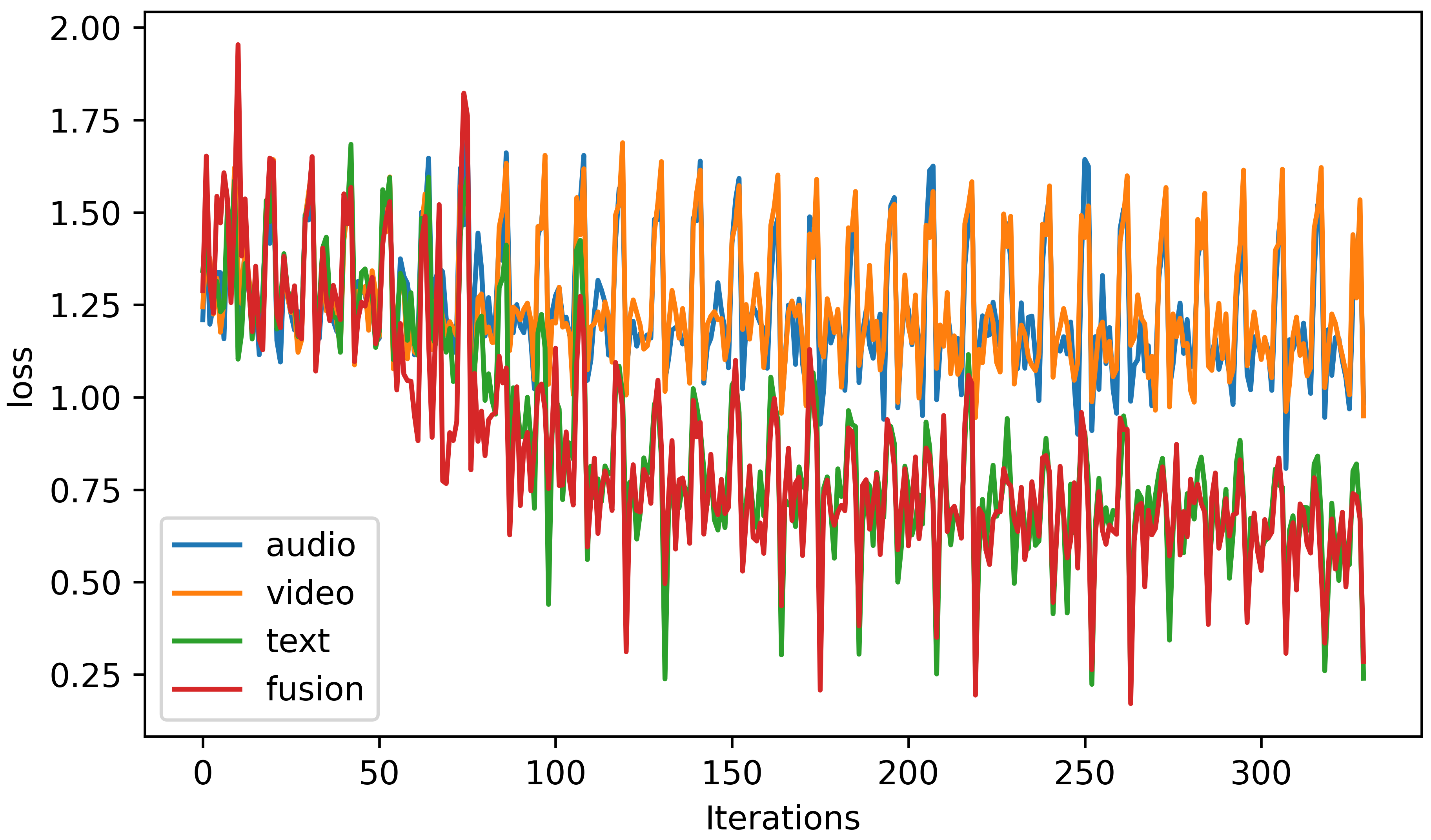}}\hspace{5pt}
 \subfigure[CGGM]{\includegraphics[width=.43\columnwidth]{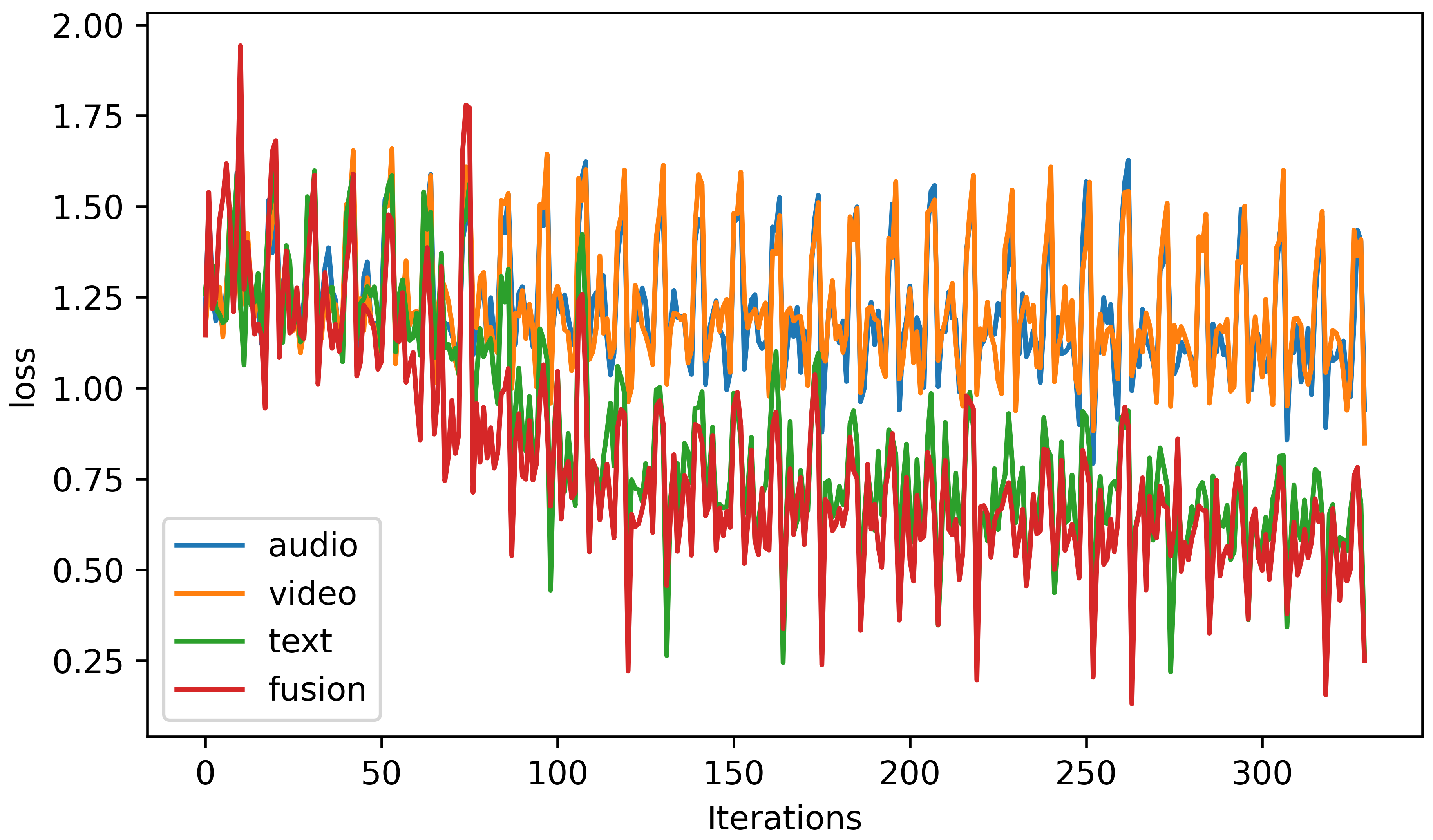}}\\
  \caption{Changes in loss during the training process.}
  \label{ab:loss}
  \end{center}
\end{figure}

\begin{figure*}
  \begin{center}
  \centerline{\includegraphics[width=0.6\columnwidth]{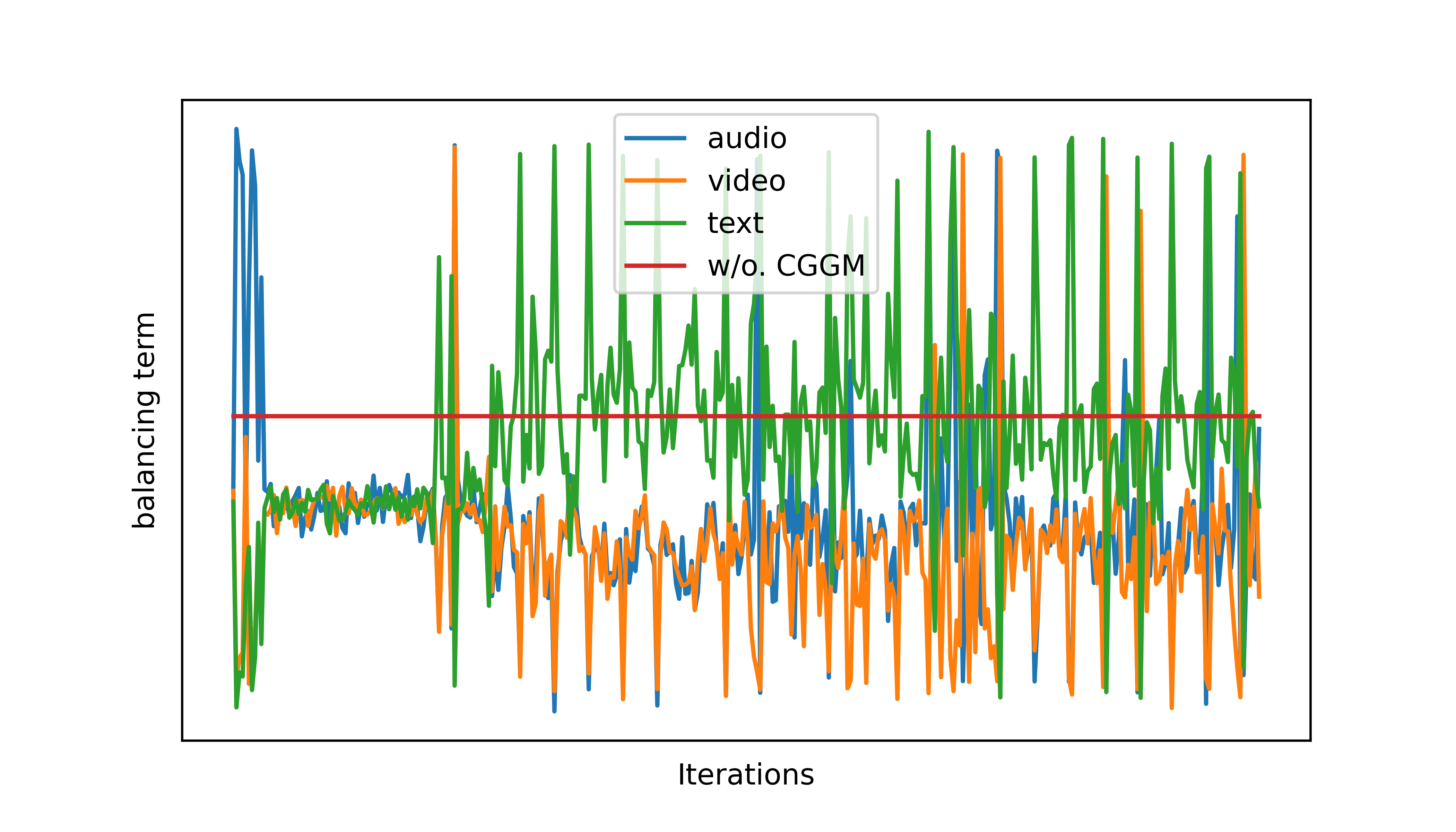}}
  \caption{Changes in balancing term during the training process.}
  \label{ab::bal}
  \end{center}
\end{figure*}

\section{More Ablation Study}\label{ap4}
\noindent\textbf{More visualizations of CGGM.} We further visualize the loss changes in Figure~\ref{ab:loss}. From the figure, we can observe that the loss of the dominant modality with CGGM implemented in (b)-(d) will drop much slower than that in Figure~\ref{ab:loss}(a). Besides, the losses of all modalities in (b)-(d) are smaller than those in (a), indicating the effectiveness of CGGM. Apart from the loss changes, we also visualize the changes in balancing term during the training process in Figure~\ref{ab::bal}. When the value is higher than the red line, the modality is promoted. When the value is lower than the red line, the modality is suppressed. In the first few iterations, the dominant modality is suppressed, ensuring that other modalities are fully optimized. During the optimization, balancing terms of three modalities turn up and down, ensuring each modality is sufficiently optimized.

\begin{table}[]
  \centering
  \caption{Additional gpu memory cost (MB) of classifiers.}
  \label{ab::com}
  \resizebox{0.65\columnwidth}{!}{%
  \begin{tabular}{@{}ccccc@{}}
  \toprule
  Setting & Food101 & MOSI & IEMOCAP & BraTS \\ \midrule
  With classifiers & +8MB & +8MB & +8MB & +24MB \\ \bottomrule
  \end{tabular}%
  }
\end{table}

\noindent\textbf{Additional computational resources of classifiers.} The additional classifiers will need more computational resources during training. However, during inference, the classifiers will be discarded. Therefore, they have no impact during the inference stage. We report the additional memory cost (MB) of the additional classifiers in Table~\ref{ab::com}. From the table, we can observe that the additional computational increase is low. There are two main reasons: (1) the classifiers or decoders are light with only a few parameters; (2) the classifiers only use the gradients to update themselves and do not pass the gradients to the modality encoders during backpropagation. Therefore, there is no need to store the gradient for each parameter, thus reducing memory cost.

\end{document}